\documentclass[a4paper,fleqn]{cas-sc}

\usepackage[numbers]{natbib}
\usepackage{booktabs}
\usepackage{array} 
\usepackage{float}
\def\tsc#1{\csdef{#1}{\textsc{\lowercase{#1}}\xspace}}
\tsc{WGM}
\tsc{QE}

\begin{document}
\let\WriteBookmarks\relax
\def\floatpagepagefraction{1}
\def\textpagefraction{.001}

\shorttitle{Binocular Model: A deep learning solution for online melt pool temperature analysis using dual-wavelength Imaging Pyrometry}    

\shortauthors{Akhavan et al.}  

\title [mode = title]{Binocular Model: A deep learning solution for online melt pool temperature analysis using dual-wavelength Imaging Pyrometry}

\author[1]{Javid Akhavan}[orcid=0000-0002-6485-5986]
\cormark[1]
\ead{jakhavan@stevens.edu}
\credit{Data Curation, Methodology, Software, AI-Modeling, Conceptualization, Investigation, Visualization, Writing Original Draft, Validation, Formal Analysis,  Writing, Review and Editing.}
\affiliation[1]{organization={Stevens Institute of Technology},
            addressline={1 Castle Point Terrace}, 
            city={Hoboken},
            postcode={07030}, 
            state={New Jersey},
            country={USA}}

\author[1]{Chaitanya Krishna Vallabh}[orcid=0000-0003-4396-3176]
\ead{cvallabh@stevens.edu}
\credit{Dataset Access, Methodology, Writing Final Draft, Review, Supervision}

\author[2]{Xiayun Zhao}[orcid=0000-0002-6092-1130]
\affiliation[2]{organization={University of Pittsburgh},
            addressline={4200 Fifth Ave}, 
            city={Pittsburgh},
            postcode={15261}, 
            state={Pennsylvania},
            country={USA}}
\ead{xiayun.zhao@pitt.edu}
\credit{Review, Supervision, Project Administration}
\author[1]{Souran Manoochehri}[orcid=0000-0002-7189-6356]
\ead{smanooch@stevens.edu}
\credit{Review, Supervision, Project Administration}


\begin{abstract}
In metal Additive Manufacturing (AM), monitoring the temperature of the Melt Pool (MP) is crucial for ensuring part quality, process stability, defect prevention, and overall process optimization. Traditional methods, are slow to converge and require extensive manual effort to translate data into actionable insights, rendering them impractical for real-time monitoring and control. To address this challenge, we propose an Artificial Intelligence (AI)-based solution aimed at reducing manual data processing reliance and improving the efficiency of transitioning from data to insight. In our study, we utilize a dataset comprising dual-wavelength real-time process monitoring data and corresponding temperature maps. We introduce a deep learning model called the "Binocular model," which exploits dual input observations to perform a precise analysis of melt pool temperature in Laser Powder Bed Fusion (L-PBF). Through advanced deep learning techniques, we seamlessly convert raw data into temperature maps, significantly streamlining the process and enabling batch processing at a rate of up to 750 frames per second, approximately 1000 times faster than conventional methods. Our Binocular model achieves high accuracy in temperature estimation, evidenced by a 0.95 R-squared score, while simultaneously enhancing processing efficiency by a factor of $\sim1000x$ times. This model directly addresses the challenge of real-time melt pool temperature monitoring and offers insights into the encountered constraints and the benefits of our Deep Learning-based approach. By combining efficiency and precision, our work contributes to the advancement of temperature monitoring in L-PBF, thus driving progress in the field of metal AM.
\end{abstract}


\begin{keywords}
Computer Vision\sep Deep Learning \sep Smart Additive Manufacturing\sep Real-Time Melt Pool Monitoring\sep Two Wavelength Imaging Pyrometry 
\end{keywords}

\maketitle

\section{Introduction} 
\label{sec:intro}

The temperature of the MP in metal AM plays a pivotal role in determining the quality and properties of the printed components. Monitoring and controlling MP temperature is essential for achieving precise and reliable results in AM processes. However, traditional MP temperature monitoring methods have long been plagued by challenges such as time complexity, manual data processing, and model development, which impede their effectiveness for time-sensitive and flexible noise-resistant tasks.
In recent years, the advent of Artificial Intelligence and Machine Learning has opened new avenues for enhancing process monitoring in AM. These advanced technologies have the potential to revolutionize the way we observe, analyze, and control the MP temperature. Among the various approaches available, AI-enabled Computer Vision stands out as a promising solution for addressing the complexities of MP temperature monitoring.
In this paper, we delve into the significance of MP temperature monitoring in metal AM, shedding light on its critical role in ensuring the printed components' quality. We review the existing state-of-the-art methods for MP monitoring, particularly focusing on the traditional arithmetic solutions that have been employed. We then explore the advantages offered by AI/ML techniques, with a particular emphasis on the reasons behind our choice of AI-enabled computer vision as the key enabler for our research.
This paper presents a comprehensive investigation into AI-driven computer vision usage for real-time MP temperature analysis. We propose a novel deep-learning model that leverages dual-wavelength photometric observations to optimize the monitoring process. our model is named binocular by getting inspired by a binocular dual input structure. This model not only overcomes the limitations of conventional methods and algorithms but also offers substantial improvements in efficiency and accuracy. In the following sections, we detail the methodology, experiments, and results that underpin this creative approach, paving the way for advancements in metal AM.

\section{Related Work}
\label{sec:related}
MP dynamics play a pivotal role in part properties analysis in Laser Powder Bed Fusion (L-PBF) based AM processes \cite{ref1,ref2,ref3}.
The properties, including fabrication quality and part mechanical properties, can be predicted based on the MP morphology, temperature profiles, and temperature transitions. However, the metal AM process's rapid transitioning and complex dynamics, present a considerable challenge to acquire immediate, real-time monitoring during the manufacturing process. The majority of proposed solutions to observe and monitor the MP dynamics in metal AM are suffering from cost-intensive or labor-intensive tools for in-situ precise observation. 

Understanding the fundamental physics governing temperature-related phenomena is crucial for analyzing, understanding, and monitoring the part properties. This involves thorough in-situ process monitoring, particularly focusing on temperature measurement, within metal AM. The heating and cooling rates significantly affect the microstructure, and the rapid laser melting and solidification process occurs within a short timeframe, typically between 200 to 600 microseconds \cite{cooling_rate}. Furthermore, the plasma generated during the process can cause reflection or refraction \cite{reflection}, which greatly affects radiation and complicates temperature measurements. Therefore, to overcome these challenges and limitations, there's a need for high-speed measurement tools that operate on a microsecond scale to capture such fast-occurring phenomena. 

 Numerous strategies have been employed to address the inherent inefficiencies and attain optimal MP dynamics. These strategies leverage a diverse array of sensory modules, with camera-based imaging emerging as a focal point due to its provision of tangible data and straightforward setup. Typically deployed for real-time process monitoring, camera-based monitoring methods also find utility in offline scenarios for capturing vital quality metrics. This methodology hinges on the acquisition of a sequence of images during the fabrication process, offering versatility in application based on the chosen light source. Notably, visible light photogrammetry \cite{MSEC_Javid} stands out as a widely favored technique, utilizing visible light transmissions to document phenomena. Conversely, infrared and multi-wavelength photogrammetry \cite{infrared} present a more intricate approach, relying on infrared emissions to generate temperature maps. Meanwhile, the utilization of X-ray \cite{X_ray} photogrammetry, although costly, delivers insights into part densities and mass distribution through external X-ray radiation.
 
X-ray-based imaging and visible light imaging methods are the most rewarding techniques utilized for in-situ monitoring of metal AM processes, each presenting its own set of challenges. For example, x-ray-based methods are most suited for revealing keyhole formation and within-part material transformations during the metal melting process in LPBF \cite{ref4,ref5,ref6,ref7}, especially for anomalies and dynamics that happen inside the part, unobservable by visible or infrared emission. However, they are concurrently hindered by limited accessibility for most AM researchers due to their complexity, safety precautions, and extreme cost. 

Visible light photogrammetry methods, on the one hand,  are relatively cheaper and easier to integrate, and also prove effective in visualizing some of the AM process dynamic aspects, such as MP morphology formations and powder motion. This effectiveness stems from the camera's ability to observe visual factors such as MP temperature map transitions and phase transformation,  microstructure development, and arrangement \cite{ref14,ref15,ref16}, residual stress flaws caused by thermal residuals, and temperature-dependent recoil pressure \cite{ref12,ref13}. On the other hand, due to the observation aspect limitations, camera-based solutions face limitations because of their limited line of view, and obstruction caused by fabrication-related anomalies such as spatters, sparks, smoke, reflections, and powder spreading \cite{ref8,ref9,ref10,ref11}. These limitations cause limited reliability and flexibility.

To enumerate several sensors utilized for process monitoring, \cite{ref21} recommended employing a commercial multi-wavelength pyrometer for monitoring electron-beam metal AM. In subsequent studies, researchers \cite{ref24,ref25,ref26}, devised co-axial monitoring systems consisting of a high-speed CMOS camera and a photodiode to capture MP radiation. Further advancements were made by authors in \cite{ref27,ref28}, who developed two-wavelength pyrometry systems with two photodiodes to capture surface thermal radiation and estimate MP temperature in LPBF AM. Expanding on this, Hooper \cite{ref29}, introduced a two-color high-speed system (100,000 frames per second) inspired by the works of \cite{ref30,ref31}, building upon the two-wavelength imaging pyrometry (TWIP) method. Alternatively, \cite{ref10} estimated MP temperature levels utilizing a single-wavelength MP emission radiance obtained from the coaxial camera, offering a different perspective on temperature monitoring. Researchers in \cite{ref32}, utilized a single high-speed camera integrated with infrared (IR) camera sensors to estimate temperature measurements and cooling rates based on emissivity values, operating at a speed of 10,000 frames per second.

Each of the approaches mentioned above exhibits significant promise alongside its distinctive limitations. Authors in \cite{ref21} demonstrate commendable accuracy; however, it is constrained to measuring the preheat temperature of the powder bed rather than the MP, particularly at slower speeds (0.125–23 Hz). The method proposed by \cite{ref24,ref25,ref26} integrates radiation from two sensors, representing MP intensity and enabling estimation of MP shape, process defects (such as balling and pore formation), and MP temperature. Nonetheless, a drawback of this system is its limited data acquisition speed, capped at 10 kHz. Building upon the work of \cite{ref27}, the authors in \cite{ref28} enhanced the original system by employing two InGaAs photodiodes, doubling the field of view from 560 µm to 1120 µm. Although offering a commendable temporal resolution of the MP temperature, these systems are confined to a fixed region of interest (a single-point measurement) and cannot provide a spatial profile. The authors in \cite{ref29}, improved both hardware and software capabilities in comparison to \cite{ref30,ref31}, resulting in enhanced frame-rate data acquisition and enabling practical application in evaluating MP heating and cooling rates using rectangular test coupons. However, the limitation remains in the restricted recording time not capable of observing a whole print.

While these methods provide valuable insights into metal AM and MP temperatures, conventional photodiodes, thermocouples, infrared cameras, and pyrometers have limitations. These limitations include speed, duration, online data capabilities, transfer rates, computational power, and cost. These factors present significant barriers to their widespread adoption for research and process control in metal AM.

The study by \cite{Base_line_paper} addressed critical gaps in in-situ MP temperature measurement and monitoring through the introduction of a novel Single-camera Two-Wavelength Imaging Pyrometry (STWIP) system. This pioneering approach leverages a single high-speed camera, thereby reducing equipment costs and the size of real-time image data. Consequently, it effectively mitigates challenges associated with image storage and transfer capabilities, enabling extended monitoring periods. Despite the notable improvement in data acquisition, boasting an impressive flow rate that could exceed 30,000 frames per second, the researchers encountered limitations in processing raw observations to derive meaningful MP signatures in real time. Their proposed method relied on the KAZE \cite{kaze} algorithm to integrate the two-wavelength observations, with each frame requiring approximately seconds to minutes to converge, and with no guarantee of convergence due to the stochastic nature of the print. Random noises such as spatter, sparks, and such would cause the algorithm to produce inaccurate key points and descriptors, potentially leading to decreased performance in feature detection and matching tasks.

To address this bottleneck, leveraging ML methods presents a promising avenue. By employing ML techniques, it becomes feasible to expedite the transformation of raw data into MP information through an end-to-end model. Building on this notion, \cite{TDIP} introduced a tunable deep image processing approach. This innovative method automates raw data pre-processing and insight generation, eliminating the need for manual intervention and offering a comprehensive end-to-end solution to enhance the efficiency and effectiveness of MP measurement and monitoring.

In this study, we extend upon the foundational methodology for MP observation outlined in \cite{Base_line_paper}, aiming to streamline data processing and eradicate the need for manual intervention in data preparation. Inspired by the innovative approach introduced in \cite{TDIP}, we present a deep learning-based solution, dubbed the binocular model, in homage to its dual observatory structure. This model offers an end-to-end framework, drastically reducing data processing time by a factor of 1000 compared to the baseline methodology. To develop this model, we leverage the dataset utilized in the baseline paper, employing rigorous modeling techniques and optimization strategies during training. The resulting model achieves an impressive data processing rate of approximately 700 frames per second, accompanied by a promising R-squared value of 0.95, signifying its robust performance and potential for advancing MP observation methodologies.

\label{sec:exp}
\section{Experimental Study for Investigation of Melt Pool Temperature Measurements}

\begin{figure}[h]
  \centering
  \includegraphics[width=0.8\linewidth]{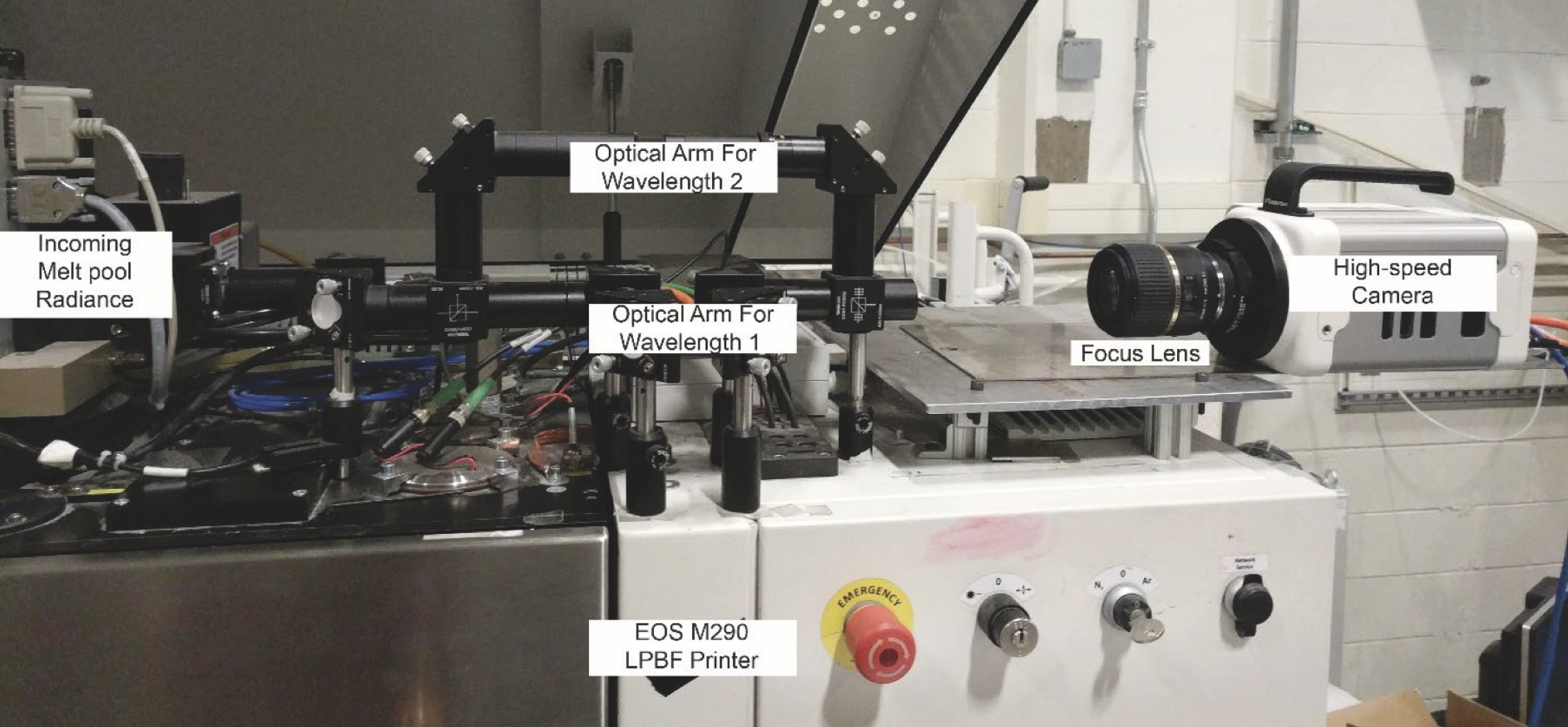}
  \caption{The experimental setup used for single-camera two-wavelength imaging pyrometry (STWIP). \cite{Base_line_paper}}
  \label{fig:experimental_setup}
\end{figure}

The authors in \cite{Base_line_paper} utilized an L-PBF (EOS M290 DMLS) AM machine for artifact fabrication and dataset generation. This machine was equipped with an in-house developed high-speed single-camera two-wavelength imaging pyrometry (STWIP) system, facilitating real-time monitoring of the fabrication process. To capture two specific wavelength observations, a dual-wavelength pyrometry setup affixed to the camera was used. This setup comprises achromatic doublets (Thorlabs), wavelength-specific band-pass filters (Thorlabs), beam-directing optics (Thorlabs), and a high-speed camera (Nova S12). The printer and camera configuration is illustrated in Figure \ref{fig:experimental_setup}. The details of the machine setup can be found in the baseline paper \cite{Base_line_paper}.

Utilizing the refined optics setup, MP data is captured at two distinct wavelengths 550nm ($\lambda_{1}$) and 620nm ($\lambda_{2}$) wavelengths in each pyrometer branch throughout the printing procedure. These wavelengths were selected based on the optimum wavelength-temperature relation and the Wien approximation. Images are acquired at a rapid rate of 30,000 frames per second and a resolution of 128 × 48 pixels, with each pixel corresponding to an optical resolution of 20$\mu m$.

\begin{figure*}
    \centering
    \includegraphics[width=0.8\linewidth]{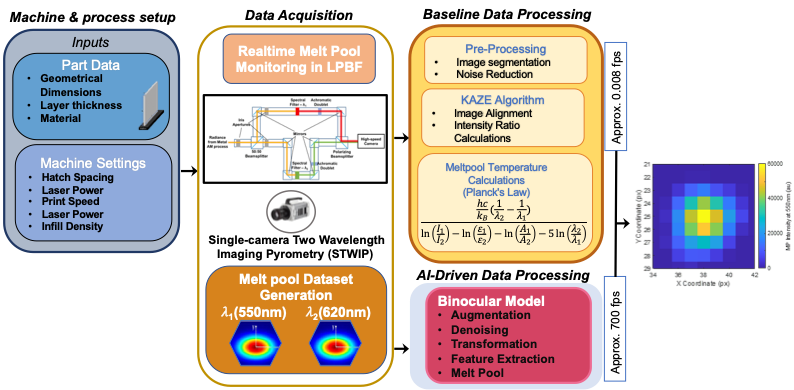}
    \caption{In this experiment flowchart, the process initiates with specimen design and fabrication plan generation on the left. Proceeding onward, fabrication and data acquisition phases ensue, culminating in comprehensive data processing. Finally, the acquired data undergoes two analyses, one using the baseline and one using AI, leading to precise estimations of the final MP characteristics.}
    \label{fig:exp_flow}
\end{figure*}

The experiment flowchart, including the baseline data computation steps and machine learning solution, is depicted in Figure \ref{fig:exp_flow}. 
\label{sec:data}
\section{Melt Pool Dataset Construction and Baseline Result Creation}


Using the developed STWIP system (shown in Figure \ref{fig:experimental_setup}), the MP data at two wavelengths (550nm and 620nm) are acquired during the print process. These MP images correspond to a fatigue test specimen print. The raw data was acquired at 30,000 frames per second with a resolution of $128\times48$ pixels at an optical resolution of $20\,\mu\text{m/pixel}$. A total of 50 print layers of data was acquired with each layer’s printing time being 72 seconds, resulting in approximately 2.2 million frames per print layer \cite{ck_1}. Testing data was also acquired in the same manner as reported in another work by the authors \cite{Base_line_paper}. The system’s optical path was validated for optical path efficiency, which is necessary for obtaining accurate temperature measurements. 

The MP temperature measurements are calculated from the intensity ratio of the MP images. The intensity ratio of the MP images is substituted into the modified Planck’s equation with Wein’s approximation for estimating the MP temperature. For obtaining accurate temperature maps, the acquired images need to be transformed accurately. The original MP images have noise and show both MPs together in one image. These images need to be split and cleaned before more processing can happen. The original images are divided into parts using common image processing methods (adaptive threshold, masking, and contouring) in MATLAB. Following this, the image registration is performed on the image for obtaining the intensity ratio measurements. 
Since both the images are imaged on the same sensor, due to their inherent wavelengths, there’s a misalignment in the scaling and rotation of the images concerning another. To obtain the accurate intensity ratio, both the MP images must be in the same spatial domain with their respective pixel intensities. To rectify this misalignment and to help digitally superpose these images, certain scaling and transformation algorithms need to be applied. The authors in \cite{ck_1} use a feature recognition algorithm, known as the KAZE algorithm. KAZE algorithm is a non-linear scale-space-based method, which is invariant to scale, and rotation, and is more distinctive at varying scales as opposed to other existing feature recognition algorithms \cite{ck_3,ck_4}. The authors implemented KAZE algorithm in MATLAB. The details about the implementation can be found in the reference article. The accuracy of the methods was also validated using similarity index functions. However, this algorithm could not be applied to a parallelized framework, since singularity solutions would break the loop and result in failed transformations. The authors report a success rate of $75\%$, which needs to be improved both in terms of accuracy and speed. Further, It has to be noted that applying the KAZE algorithm to noisy images can lead to very large errors in misalignment, which will lead to impractical MP temperatures. Therefore the pre-processed images need to be noise-free. This work aims to bridge the gap between the discussed accuracy and speed along with the minimization of noise-related errors.
It must be noted that the intensity, scaling, and rotation between the two-wavelength MP images are subject to change throughout an LBPF-based AM process due to the print process parameters, hatching patterns, and other stochastic disturbances \cite{ck_1}. Therefore, constant scaling and rotation factors cannot be assumed. Every pair of the MP images need to be processed for their specific transformation criteria for successfully estimating the temperature profile.

A representative pair of MP images transformed using the KAZE algorithm are shown in the following figure \ref{fig:MP_side_by_side}. These images are represented in a color scale to illustrate the correlation between them. It's worth noting that grayscale images wouldn't adequately depict how the two wavelength images are oriented or aligned. Following the transformation, both the X- and Y-coordinates remain consistent for the two-wavelength images, which simplifies the process of measuring intensity ratios on a pixel-by-pixel basis. The blue background in these images signifies areas devoid of MP pixels and serves as a representation of pseudo-zero intensity pixels.

\begin{figure}[h]
    \centering
    \includegraphics[width=0.8\linewidth]{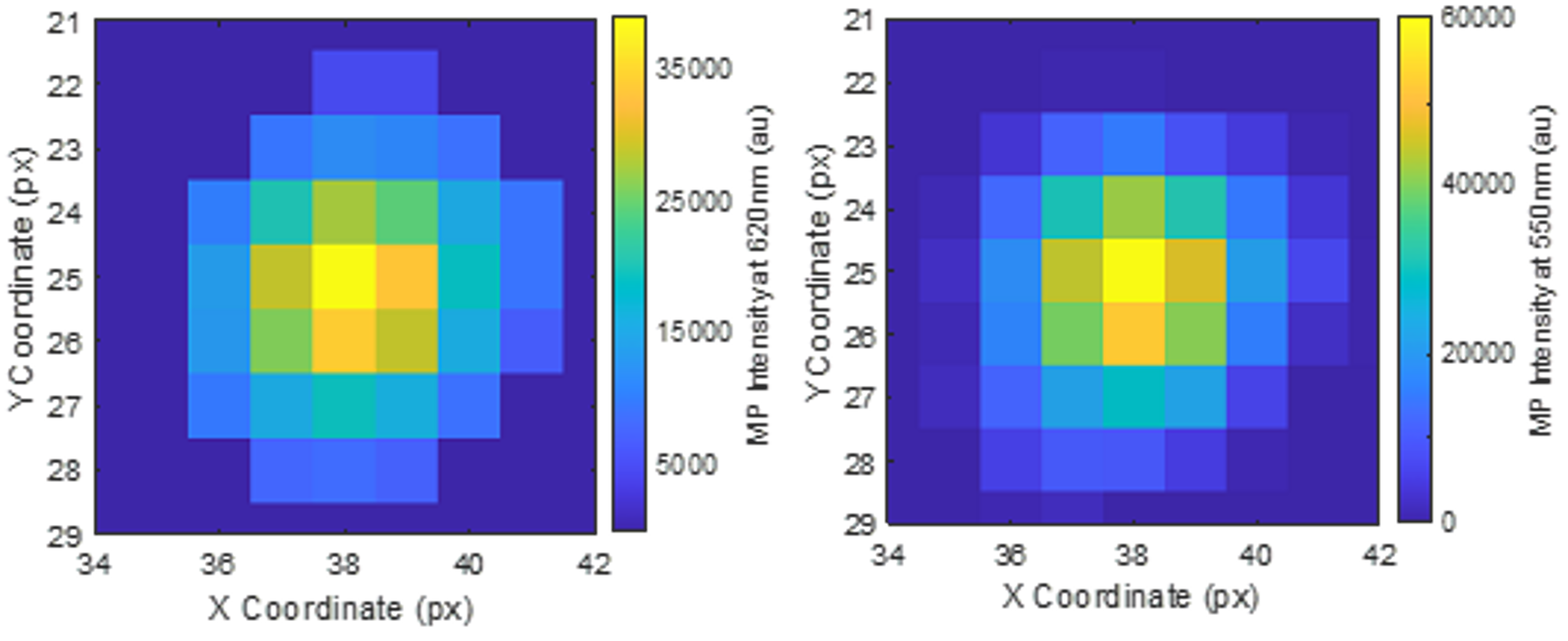}
    \caption{The illustration displays two images side by side: the original MP image at 620nm on the left and the transformed image at 550nm on the right.  \cite{ck_1}}
    \label{fig:MP_side_by_side}
\end{figure}


\label{sec:problem}
\section{Proposed AI-driven Model Design and Structure}
Starting from the inputs, the two frames undergo a pre-processing step to eliminate noise, align MPs, and share the same coordinates. As elaborated in the baseline methodology \cite{Base_line_paper}, this process requires a feature detection and description algorithm such as the KAZE algorithm to determine the appropriate rotation and alignment for each frame in the database. However, reliance on such methods introduces uncertainties in real-time processing and data-to-insight transformation. Instances may arise where the KAZE algorithm fails to converge due to the stochastic environment noise or requires longer processing times exceeding minutes per frame making it not suitable for immediate real-time tasks, as discussed in the baseline paper \cite{Base_line_paper}. 

Consequently, adopting a deep model-based solution offers significant advantages to this MP monitoring approach. Through extensive training, the model learns the necessary transformations and can generate results irrespective of singularities observed in the input, ensuring a more reliable and efficient data-to-insight transformation process.

The study's objectives revolve around two key aims: 1) Streamlining analysis through automation to minimize manual filter design and processing, also 2) Improving processing efficiency to enable batch processing and achieve near real-time data-to-insight conversion. In pursuit of these goals, a deep learning model has been conceptualized and deployed. For the model's design and implementation, the TensorFlow V2.10.0 library and NVIDIA RTX A6000, equipped with 48 GB of GPU memory, serve as the processing unit.

\begin{figure*}[ht]
  \centering
  \includegraphics[width=.8\linewidth]{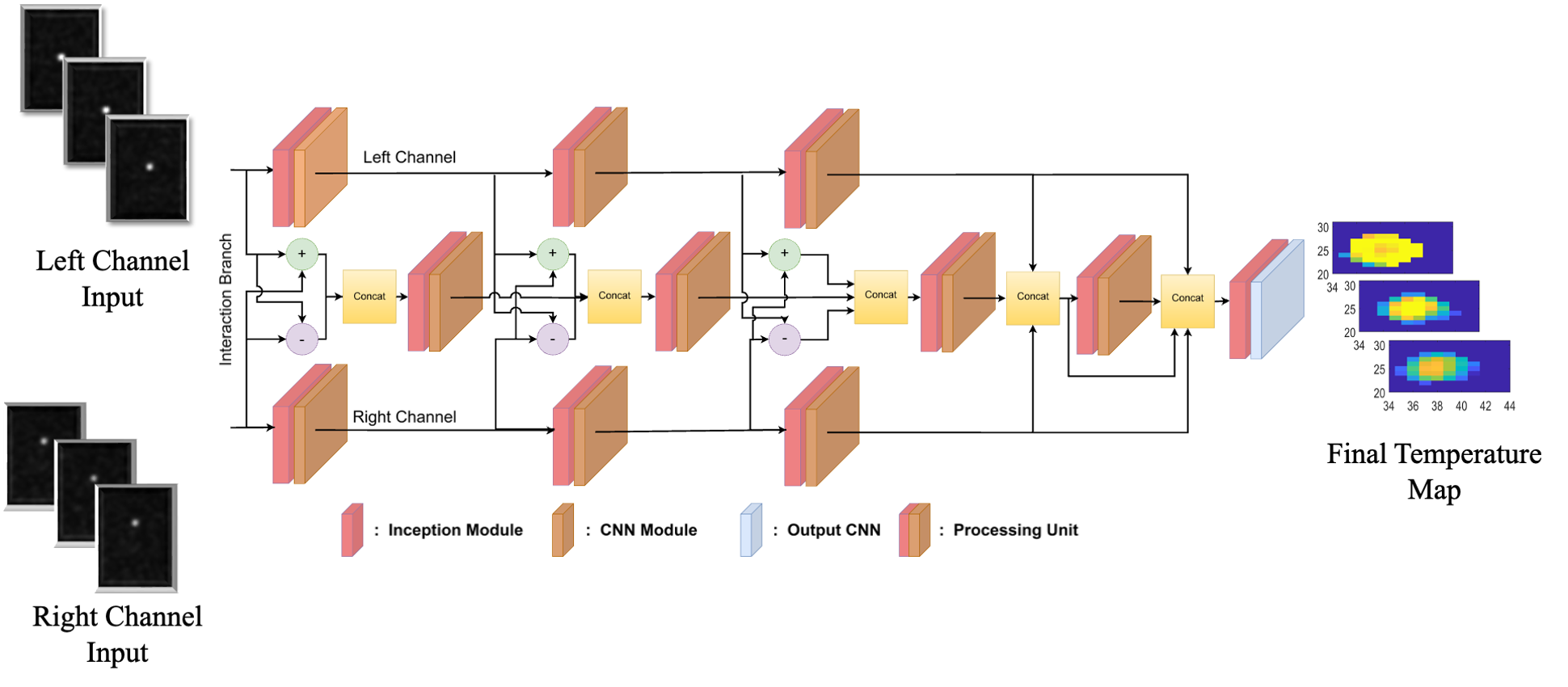}
  \caption{Binocular model structure. On the left, two-channel input images are the inputs transitioned to their corresponding MP representations on the right}
  \label{fig:Binocular_model}
\end{figure*}

The model envisioned for this task is anticipated to intake two wavelength observations without preprocessing and generate a temperature map utilizing the dual-wavelength intensity ratio map methodology detailed in the baseline methodology \cite{Base_line_paper}.
The Binocular model architecture is illustrated in Figure \ref{fig:Binocular_model}, drawing inspiration from the baseline method and the steps involved in calculating the MP temperature maps. Deep learning models, given sufficient training, possess the capability to replicate the intricate procedures and computations required to transform inputs into desired outputs. The designed model structure and mid-stage data expectation enforced with the labels can lead the network to replicate explainable procedures. 

Harnessing the computational efficiency of tensor-based operations in GPU-based methods such as AI-enabled models, allows data to be processed in batches leading to a significant computational efficiency. Consequently, instead of computing a single MP temperature map, multiple MP temperature map frames can be generated simultaneously. This approach substantially reduces processing time by a $1000\times$ factor, facilitating a more viable real-time transition from data to insights. The number of batches processed concurrently depends on the GPU's memory accessible to the model.

\subsection{Binocular Model}
To replicate the baseline \cite{Base_line_paper} approach, the KAZE algorithm, we devised a model inspired by the concept of binocular vision, employing two channels termed the left channel and the right channel. This model takes two inputs, with each wavelength data fed into a separate branch, thereby creating two distinct data branches akin to a binocular setup. The interaction between these channels is crucial to calculate intensity ratio maps. Hence, after each processing step, a data merge operation ensues, yielding a third branch known as the Interaction Branch.

The Interaction Branch begins by comparing the two input frames, performing addition and subtraction operations, and retaining the raw data of each frame. The addition operation accentuates the shared characteristics, whereas the subtraction operation highlights the disparities. These operations are fundamental for discerning both similarities and differences between the two input channels. The resultant data are concatenated and fed into a Wide Deep Module (WDM) that acts as a Processing Unit. Each branch employs a Wide-Deep network model, as described in the following section, to process and analyze the data flow.

Following each Processing Unit, data sharing, skip connections, and feed-forward mechanisms come into play. Data sharing ensures that the results of each stage remain accessible to the mid-branch, while skip connections are implemented to facilitate gradient propagation through deeper layers. Each processing unit can be considered a segment along with its associated data contribution. Stacking four such segments forms the data-processing and merge section. The output of this section is expected to closely align with the intensity ratio map generated by the KAZE algorithm discussed in the baseline paper \cite{Base_line_paper}.

To attain the anticipated MP, the output of the merged section undergoes two additional Processing Units and is subsequently mapped to a $2D$ output corresponding to the MP.

\subsubsection{Wide Deep Module Description}
\begin{figure}[h]
  \centering
  \includegraphics[width=0.8\linewidth]{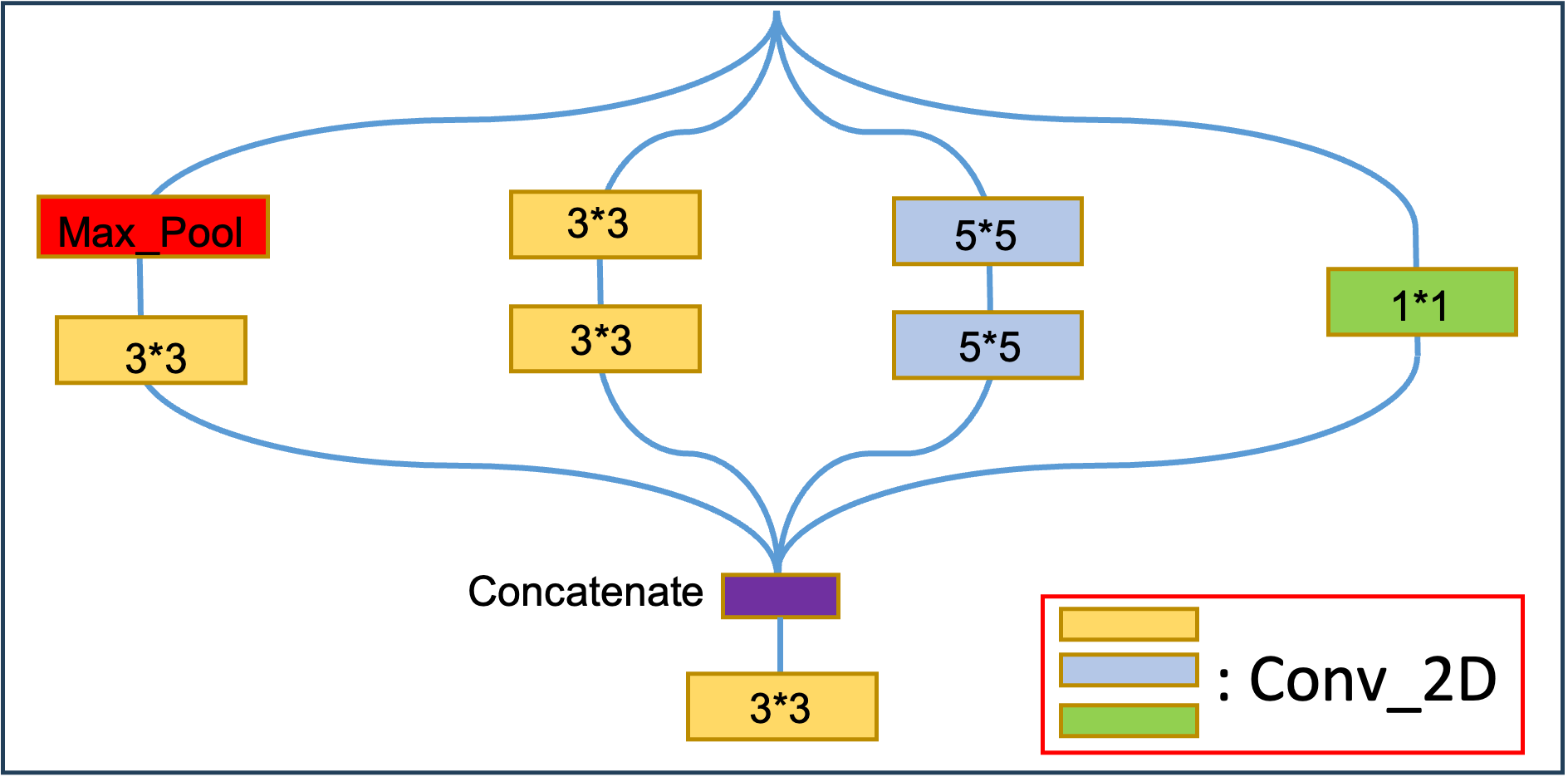}
  \caption{Wide Deep module structure, branching the data into four flows and aggregating them back for a diverse feature development}
  \label{fig:Inception_}
\end{figure}

We introduce the Wide Deep Module (WDM) as a pivotal component in our deep learning methodology as depicted in  \cite{Inception}. This module comprises an Inception module followed by a $2D$ Convolutional Neural Network (CNN) block, meticulously crafted for efficient feature extraction. The Inception module, as depicted in Figure \ref{fig:Inception_}, expands the data into four branches, enabling the exploration and analysis of a diverse array of filters and features. The multi-branch structure of the Inception module aids in mitigating gradient loss, particularly as the model scales in complexity. By consolidating the outputs of all four branches into a single tensor, the Inception module yields a thorough representation of its input data.

Subsequently, to curtail the data dimensionality, a CNN block is stacked atop the Inception module concatenate output, regulating and compressing the data flow dimension. The WDM serves as the core processing unit within the Binocular model, facilitating efficient information processing and feature extraction.

\subsection{Binocular Model Training}
\begin{figure}[h]
  \centering
  \includegraphics[width=0.8\linewidth]{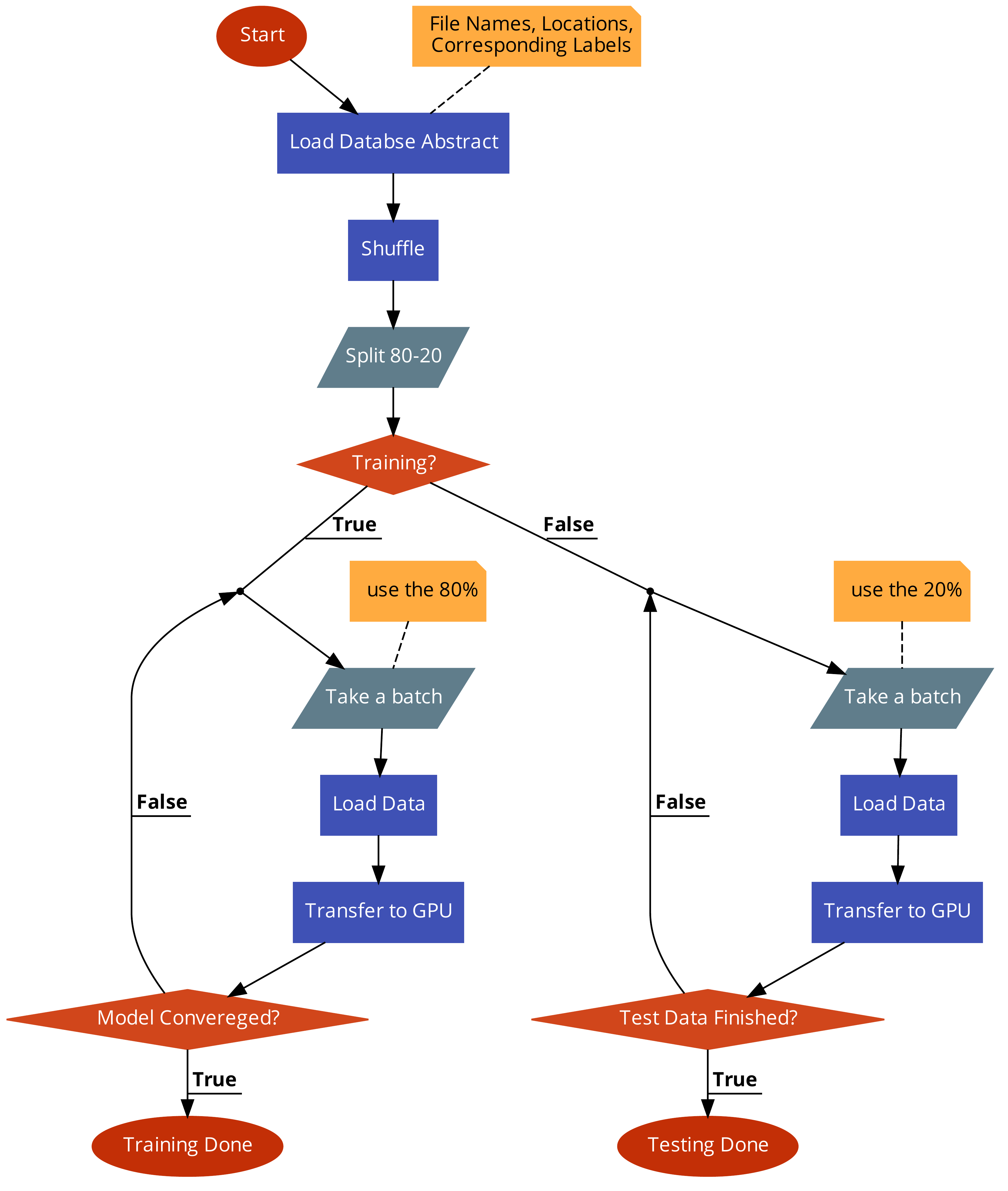}
  \caption{MP dataset data flow diagram}
  \label{fig:data_flow}
\end{figure}

To facilitate training of the Binocular model, a data flow method, illustrated in Figure \ref{fig:data_flow}, is implemented to extract images from the disk and transmit them along with their corresponding labels to the GPU. Given the large size of the database surpassing available RAM, image locations are cataloged alongside their labels to establish a reference table instead of storing full image data. Subsequently, employing an $80/20$ ratio, training, and test data are randomly partitioned to form datasets. When specific image data is needed, the data loader module retrieves the image location and loads both the raw input and label information into RAM using the recorded addresses. The data loader's batch size is optimized to maximize performance while making efficient use of available RAM capacity.

\begin{figure}[h]
  \centering
  \includegraphics[width=0.8\linewidth]{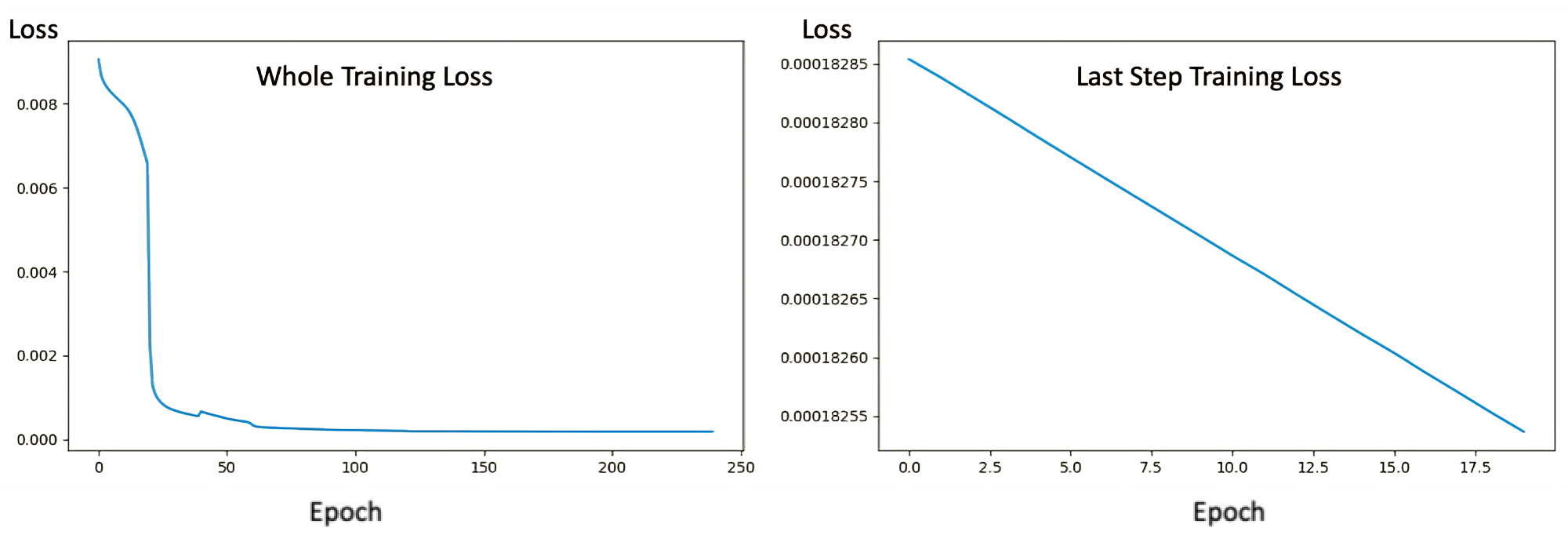}
  \caption{Training loss illustration. Left: Whole training loss history, Right: Last 20 epochs}
  \label{fig:Training_loss}
\end{figure}

The model's training strategy involves dynamic parameter updates aimed at guiding the model towards convergence from local minima to as close as the global minima. The optimizer chosen for this task was ADAM optimizer defined by \cite{ADAM_Optimizer}, and mean square error as a loss function. By employing a learning rate schedule batch sizes, and $20$ epochs per combination, the model initially grasps the data's general features. Subsequently, through fine-tuning the learning rate and reduction of batch size, the model delves deeper into the feature space, capturing more intricate details. Figure \ref{fig:Training_loss} illustrates the training loss progression over time for the Binocular model during the dynamic parameter update process. Table \ref{tab:lr_btch_ranges} provides a summary of the ranges utilized for model training.

\begin{table}[h]
\centering
\caption{Ranges of Training Learning Rates and Batch Sizes}
\label{tab:lr_btch_ranges}
\begin{tabular}{cccc} 
\toprule
\textbf{Learning Rate Range} & \textbf{Batch Size Range} \\ \midrule
1e-4, 1e-5, 1e-6, 1e-7 & 50, 20, 5 \\ \bottomrule
\end{tabular}
\end{table}

To mitigate the model's overfitting and improve generalization to unseen data, several precautions were implemented. Initially, the input data underwent augmentation through various transformations: random rotations ranging from $-10$ to $10$ degrees in 2-degree increments (yielding $11$ variants), mirroring and flipping (producing $8$ variants), and pixel shifts from $-8$ to $8$ pixels in 2-pixel intervals (resulting in $7$ variants). This process generated a total of $616$ possible representations for each input image, thereby ensuring the model's ability to learn from diverse perspectives without relying on specific locations or orientations.

Moreover, given the data's diversity, the model is anticipated to extract meaningful information and discern it from noise effectively. Concurrently, dropout layers \cite{dropout} were integrated to mitigate the model's reliance on specific information by randomly eliminating $20\%$ of data flow during training. Notably, dropout layers remain inactive during testing and real-world implementation, ensuring their effects are confined to the training phase.

\label{sec:Results}
\section{Binocular Model Performance Results And Validation}
Once the model training is completed and reasonable convergence is observed, assessing the model's performance becomes imperative. Given that process automation, accuracy, and efficient batch processing were the primary objectives of the Binocular model, it is crucial to evaluate its performance in these areas. Initially, the accuracy and reliability of the model's estimations were verified, followed by bench-marking processing time to determine the highest frame-per-second processing capacity.

\begin{figure*}[!h]
  \centering
  \includegraphics[width=0.8\linewidth]{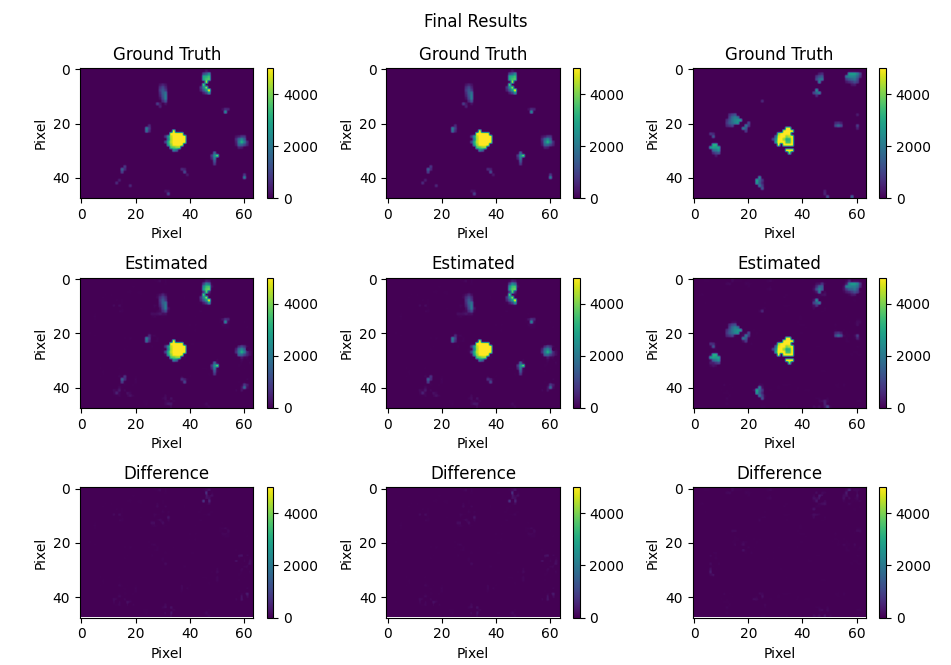}
  \caption{Three sample ground truth and Binocular model prediction. The top row shows the ground truth,  the middle is the Binocular model estimated result, and the bottom is the pixel-by-pixel difference}
  \label{fig:sample_results}
\end{figure*}

To assess the Binocular model accuracy, two measures are employed. The primary approach is the R-squared criteria. This method, also known as the coefficient of determination, evaluates the efficacy of independent variables in elucidating the variability within a regression model's dependent variable. The R-squared value, a dimensionless metric ranging from 0 to 1 according to Equation \ref{eq:rsquared}, indicates the model's performance: the higher the value, the better. Utilizing the test set, temperature maps as seen in the first row of Figure \ref{fig:sample_results} are generated pixel by pixel using the developed model. Remarkably, the Binocular model yielded an R-squared value of 0.95, underscoring the reliability of its outputs.
\begin{equation}
\begin{gathered}
R^2 = 1 - \frac{{\sum_{i=1}^n (y_i - \hat{y}_i)^2}}{{\sum_{i=1}^n (y_i - \bar{y})^2}}   \\ \text{where :} \\ \hat{y}_i \text{: is the expected value for MP temperature pixel-i} \\ {y_i} \text{: is the predicted value for MP temperature pixel-i} \\ \bar{y} \text{: is the mean of all MP temperature pixels}
\label{eq:rsquared}
\end{gathered}
\end{equation}

Figure \ref{fig:sample_results} illustrates three randomly selected data points. The top row displays the anticipated temperature maps derived from the baseline method \cite{Base_line_paper} using the KAZE algorithm, while the middle row showcases the same output generated by the Binocular model. The bottom row depicts the pixel-by-pixel differences between the two outputs to showcase how close the estimations are made using the Binocular model. In particular, the model exhibits excellent noise reduction and achieves impressive temperature map estimations.

\begin{figure}[h]
  \centering
  \includegraphics[width=0.8\linewidth]{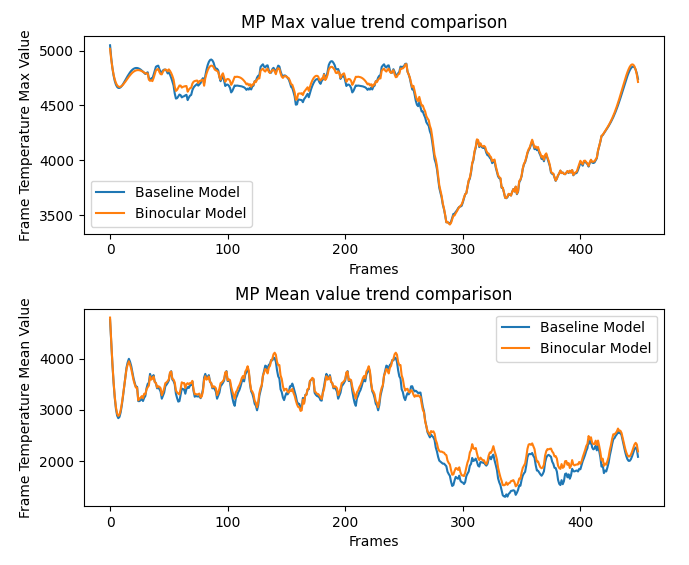}
  \caption{Comparison of Baseline Model (manual implementation of the KAZE algorithm) data vs Binocular model temperature estimation output}
  \label{fig:Mean_Max_Rep}
\end{figure}

To demonstrate the model's capabilities to follow the MP temperature transition trends and how accurate the model performs in estimating the MP, the mean and max value of various frames' MP from the baseline database \cite{Base_line_paper} and the Binocular model are calculated and compared. First, the MP area is detected by finding the maximum heat point in each data frame. Then, by growing the region around the maximum temperature point by including the surrounding pixels, the MP area is identified. To ensure a reliable and noise-free measure for the maximum value detection and calculation, the top 3 max points' average temperature is used. To calculate the mean value, the same identified MP area in the above section is averaged. The resulting maximum and average temperatures from the baseline and Binocular model are shown in Figure \ref{fig:Mean_Max_Rep}. The binocular model achieved similar results and followed the temperature trend achieved by the time-intensive baseline model.

Having a reliable prediction result and great accuracy, it is important to underscore the Binocular model's processing efficiency. In pursuit of this objective, the model's processing prowess underwent rigorous benchmarking, consistently achieving an impressive throughput ranging from 650 to 800 frames per second. It is worth mentioning that the model's architecture leverages GPU-based libraries, amplifying its computational capabilities. 

Beyond the model's promising processing time, attention must be drawn to the latency incurred during data transfer phases, encompassing disk-to-memory and memory-to-GPU exchanges, and vice versa. Through meticulous monitoring, it became evident that the frames per second performance observed in this study is constrained by the rate of data transfer between the disk, RAM, and GPU, while the GPU itself demonstrates the capacity for handling larger datasets concurrently. 
In summary, the optimization of the Binocular model's performance stands to gain significantly from enhancements in the data flow speed throughout the system. Addressing the bottleneck in data transfer holds promise for unlocking even further potential in the model's processing capabilities, thereby advancing its utility for real-time data processing. 


\label{sec:conclusion}
\section{Conclusion}
In conclusion, the evaluation of the Binocular model's performance reveals its effectiveness in achieving the primary objectives of process automation, accuracy, and efficient batch processing. Through rigorous assessment, the model demonstrates commendable accuracy, as indicated by a high R-squared value of 0.95, reflecting its reliability in MP temperature estimation. Additionally, inspection of temperature maps illustrates the model's ability to reduce noise and provide accurate estimations, as evidenced by comparison with ground truth data. Moreover, the model showcases its capability to accurately track MP transition trends, with mean and maximum temperature values closely matching those of the baseline model. 
The model's processing limits was also measured and achieved a range of 650-800 frames per second which is a significant improvement compared to the baseline approach of using KAZE algorithm taking a few seconds per frame.  
These results underscore the efficacy and reliability of the Binocular model in fulfilling its intended objectives, marking a significant advancement in temperature estimation for AM processes.

\label{sec:credits}
\section{Declaration of Competing Interest}
The authors declare that they have no known competing financial interests or personal relationships that could have appeared to influence the work reported in this paper.
\section{Declaration of Generative AI and AI-assisted Technologies in The Writing Process}
During the preparation of this work, the authors used ChatGPT in order to refine the grammar and wording. After using this tool, the authors reviewed and edited the content as needed and take full responsibility for the content of the publication.
\section{Acknowledgments}
The authors wish to extend their appreciation to Prof. Nikhil Muralidhar and Nastaran Soofi for their assistance in the manuscript review.

\printcredits

\bibliographystyle{cas-model2-names}
\bibliography{main}



\end{document}